\documentclass[
]{ceurart}

\usepackage{listings}
\newcommand{\dset}{{{ProLiFIC}}}
\newcommand{\dsetext}{Procedural Lawmaking Flow in Italian Chambers}
\newcommand{\github}{{\url{https://github.com/matildeec/ProLiFIC}}}
\newcommand{\colab}{{\url{https://colab.research.google.com/github/matildeec/ProLiFIC/blob/main/ProLiFIC_EDA.ipynb}}}
\newcommand{\code}[1]{\texttt{\scriptsize#1}}
\usepackage{xcolor}
\newcommand{\coma}[1]{{\color{red}[AV: #1]}}

\usepackage{subcaption}

\begin{document}



\conference{}

\title{
The ProLiFIC dataset: 
Leveraging LLMs to
Unveil the Italian Lawmaking Process
}



\author[1]{Matilde Contestabile}[%
email=m.contestabile@santannapisa.it
]
\address[1]{Sant'Anna School of Advanced Studies Pisa,
  Piazza Martiri delle Libertà 33, Pisa, 56127, Italy}
\address[2]{DTU Technical University of Denmark,
  Anker Engelunds Vej 1, Kongens Lyngby, 2800, Denmark}

\author[1]{Chiara Ferrara}[%
email=c.ferrara@santannapisa.it
]

\author[1]{Alberto Giovannetti}[%
email=a.giovannetti@santannapisa.it
]

\author[1]{Giovanni Parrillo}[%
email=g.parrillo@santannapisa.it
]

\author[1,2]{Andrea Vandin}[%
orcid=0000-0002-2606-7241,
email=a.vandin@santannapisa.it
]


\begin{abstract}
%
%
%
Process Mining (PM), initially developed for industrial and business contexts, has recently been applied to social
systems, including legal ones. However, PM's efficacy in the legal domain is limited by the accessibility and
quality of datasets. 
We introduce ProLiFIC (Procedural Lawmaking Flow in Italian Chambers), a comprehensive
event log of the Italian lawmaking process from 1987 to 2022. Created from unstructured data from the Normattiva
portal and structured using large language models (LLMs), ProLiFIC aligns with recent efforts in integrating PM
with LLMs. We exemplify preliminary analyses and propose ProLiFIC as a benchmark for legal PM, fostering
new developments.

\end{abstract}


\maketitle

\section{Introduction}
The Italian lawmaking process involves intricate institutional procedures, complex interactions among multiple actors from two separate chambers with equal powers (Senate and Chambers of Deputies), and numerous formal stages. Understanding the legislative pathways is crucial for institutional transparency and empirical research in law, political science, and public administration. 
Previous research focused on aspects like legislative output, raw performance,  party coalitions, party switches,  and policy agendas (see, e.g.,~\cite{Kreppel1997,MENEGHETTI2023107098,Borghetto2012A}).
Despite the growing efforts to study the legal domain with process mining techniques (see, e.g.,~\cite{caponecchia2024, lopez2021}),  there is a lack of process-oriented studies focusing on 
tracking the internal procedures of lawmaking. 
We believe that this is due to the lack of machine-accessible high quality datasets on these aspects. 

The entire life-cycle of each approved Italian law is documented in unstructured textual records, the \textit{lavori preparatori} (preparatory works), collected in the Normattiva portal (\url{https://www.normattiva.it/}) handled by the Italian Presidency of the Council of Ministries. 
While publicly available, the unstructured and textual nature of preparatory works severely impedes systematic process-oriented analysis tasks. 

To address this challenge and unlock the analytical potential within these texts, we introduce ProLiFIC (Procedural Lawmaking Flow in Italian Chambers). This is an event log dataset that extracts and structures events from Normattiva's preparatory works. The dataset has been meticulously created using an automated pipeline powered by Large Language Models (LLMs), aligning with recent efforts in bridging process mining with LLMs by leveraging their advanced natural language understanding capabilities, as demonstrated in e.g. in~\cite{chalkidis2020legal}, to transform unstructured textual data into structured, actionable event logs. 
This resource represents a significant contribution to the Business Process Management (BPM) field. 
It offers great practical value by supporting process-oriented analyses that have the potential to provide actionable insights to improve the transparency and enhance the efficiency of such a crucial sector, informing policy. Specifically, just to cite a few examples, ProLiFIC may facilitate: (i) the comparative analysis of lawmaking across different legislatures; (ii) the identification of procedural bottlenecks by pinpointing 
inefficiencies in legislative workflows; and (iii) studying the impact of changes in the composition of the chambers or in the laws governing lawmaking.  In essence, ProLiFIC transforms vast amounts of unstructured texts into a powerful event log resource for empirical research, 
offering a novel perspective to analyze and improve the Italian legislative system. 
\dset{} is available on Zenodo at~\cite{prolificzenodo}.\footnote{In addition, the working GitHub repository of \dset{} is available at \github, where it can be tested online on Google Colab without requiring any local installation (see \colab{}).} 

Our work is related to the ILMA dataset~\cite{ilma2012}, used in~\cite{Borghetto2012A} to study the temporal dimension of Italian lawmaking. This is a thorough dataset, with a wider scope than \dset. E.g., it includes information on parliamentary votes, parties, etc.  
However, these studies use less fine-grained data on procedural aspects, do not use process mining, and consider only a limited time period (1987–2008)\footnote{Notably, as of this writing the project appears to be discontinued and unavailable  \url{https://159.149.130.120/ilma/sito/}.}.


\section{Characteristics of the dataset, and data model}

\dset{} covers 4395~\footnote{In data cleaning, we identified about 156 laws with factual problems (such as typos) in Normattiva. We corrected them when the intended meaning was evident from the context, while 77 laws were excluded due to unresolvable ambiguities.
} 
Italian laws enacted between 1987 and 2022 (legislatures X to XVIII). These correspond to all the complete legislatures for which preparatory works are available in the Normattiva portal.
\dset{} is a structured dataset of preparatory works of such laws that combines two complementary components: (i) metadata capturing substantive attributes of each law and (ii) an event log detailing the sequence of legislative actions taken during the law’s lifecycle. These components are linked through a shared identifier, \texttt{case\_id}, corresponding to the official \textit{codice redazionale} (editorial code) assigned upon publication in the \textit{Gazzetta Ufficiale} (a formal step which marks the enactment of the law). 

\smallskip
\noindent\textbf{Metadata Log part of \dset{}.}  
Table~\ref{tab:metadata} shows the attributes stored in the metadata. They include general information, 
such as formal characteristics and  dates. 
%
%
For example, a law may consist of multiple articles. Also, in addition to  `ordinary' laws, we may have decree conversions (urgent government
measures that must be converted into law within 60 days) or EU ratifications (the act of importing EU regulations in the Italian system). 
Furthermore, we record the legislature and the government under which each law was enacted (since Italy is a parliamentary republic, a single legislature may include multiple governments).

\smallskip
\noindent\textbf{Event Log part of \dset{}.}  
Table~\ref{tab:event_log} illustrates the structure of the event log. It captures the procedural trajectory of each law 
as a sequence of time-stamped events, all sharing the same \texttt{case\_id}. 
Each event involves an  activity, namely 
\code{Presentation} (the law is introduced in a chamber),   \code{Assignment} (each chamber has a number of topic-specific committees which look in detail at a law and de facto perform the bulk of the legislative work -- this event marks the assignment of the law to a committee), 
\code{Committee examination} (the committee assesses the law), \code{Request for opinion} (a committee may ask for further opinions from additional committees), \code{Report} (the conclusions of the committees are presented to the assembly), \code{Plenary debate} (the whole assembly discusses it), and 
\code{Approval} (the law, potentially modified in the previous steps, is approved in a chamber). 

During the works in a chamber, the law may be amended (either during \code{Committee examination} or \code{Plenary debate}). 
Although the preparatory works do not contain detailed information about the amendments themselves, when a law is approved with amendments, it must be sent back to the other chamber to approve the revision. This back-and-forth process between the two chambers, known as \emph{navetta} (\emph{shuttle}), continues until both chambers approve the exact same version of the law. Once this occurs, the preparatory works conclude, and so does tracking in Normattiva and in \dset.
%
Each event may have further specific information like the committee or person involved, and the current chamber. In order to facilitate validation and explainabilty, each event is paired with the original chunk of text that originated it. 

\begin{table}
\centering
\begin{minipage}[t]{0.45\textwidth}
\centering
\scalebox{0.67}{
\begin{tabular}{ll}
\toprule
\textbf{Attribute} & \textbf{Description} \\
\midrule
\texttt{case\_id} & Unique identifier for the law \\
\texttt{title} & Official title of the law \\
\texttt{legislature} & Legislative session number \\
\texttt{government} & Executive in office at the time \\
\texttt{publishing\_date} & Date of publication in the \textit{Gazzetta Ufficiale} \\
\texttt{implementation\_date} & Date law takes effect (15 days after publication) \\
\texttt{decree\_conversion} & Indicates if it is a decree-law conversion \\
\texttt{eu\_ratification} & Indicates if the law results from EU indications \\
\texttt{articles} & Number of articles in the law \\
\texttt{description} & Short summary of the law's purpose \\
\texttt{full\_text\_url} & Link to full legal text on Normattiva \\
\bottomrule
\end{tabular}
}
\caption{Metadata attributes for each law. One row per law.}
\label{tab:metadata}
\end{minipage}
\ \ \ \ \ 
\begin{minipage}[t]{0.45\textwidth}
\centering
\scalebox{0.67}{
\begin{tabular}{ll}
\toprule
\textbf{Attribute} & \textbf{Description} \\
\midrule
\texttt{case\_id} & Unique identifier for the law \\
\texttt{chamber} & \textit{Camera} (Chamber of deputies) or \textit{Senato} (Senate)\\
\texttt{activity\_it} & Type of legislative action (e.g., \textit{Presentazione}, \textit{Approvazione}) \\
\texttt{activity\_en} & English translation of \texttt{activity\_it} \\
\texttt{time} & Timestamp of the activity, with granularity given in days \\
\texttt{committee} & Committe involved in the event 
\\
\texttt{person} & Sponsor, speaker, or institutional actor proposing the law  
\\
\texttt{chunk} & Extract of legal text or speech related to the activity \\
\bottomrule
\end{tabular}
}
\caption{Event log with all events of all laws. One row per event.}
\label{tab:event_log}
\end{minipage}
\end{table}

 
\section{Preliminary analyses}

In this section, we provide an initial exploration of the dataset through descriptive statistics and process maps elaborated by Fluxicon Disco~\cite{discoo}, examining how legislative output, duration, and procedural flows vary across Italian legislatures.

\smallskip
\noindent\textbf{Descriptive statistics.}  
We begin by summarizing legislative activity over time. In Table~\ref{tab:leg_description_extended}, for each legislature we report the number of months it elapsed, the number of governments it included, the total number of enacted laws (including the percentage of decree-laws), and the median duration of the legislative process. The latter is calculated as the number of days elapsed between the first and last recorded event in the law’s procedural history. 
{The table does not show evident trends. 
However, we note a decrease over time in the number of laws enacted in  the legislatures. In particular, the first two legislatures (X and XI, 1987-94) involve 1363 laws over 80 months, while the last two (XVII and XVIII, 2013-22) only 677 over 114 months. At the same time, the last two legislatures show higher median durations for non-decrees (the conversion of decrees is bound by law by 60 days, and all legislatures have comparable median durations for these). Next we will see how simple process-oriented analyses can instead highlight significant changes in the two timeframes in the Italian lawmaking process.} 

\begin{table}
\centering
\scriptsize
\setlength{\tabcolsep}{11pt}
\begin{tabular}{lcc cccccc}
\toprule
\multicolumn{3}{c}{\textbf{Legislature}} & 
\multicolumn{2}{c}{\textbf{Number of laws}} &
\multicolumn{2}{c}{\textbf{Median Duration \textit{(days)}}} \\
\cmidrule(lr){1-3} \cmidrule(lr){4-5} \cmidrule(lr){6-7}
\textbf{Id} & \textbf{Length \textit{(months)}}& \textbf{Governments} & \textbf{Total} & \textbf{\%Decrees} & \textbf{Ordinary} & \textbf{Decrees} \\
\midrule
X & 57 & 4 & 1060 & 19\% & 368 & 55 \\
XI & 23 & 2 & 303 & 38\% & 271 & 55 \\
XII & 24 &2 & 292 & 43\% & 209 & 51 \\
XIII & 60 & 4 & 892 & 21\% & 343 & 54 \\
XIV & 58 & 2 & 676 & 30\% & 268 & 52 \\
XV & 24 & 1 & 110 & 29\% & 214 & 51 \\
XVI & 58 & 2 & 385 & 30\% & 161 & 54 \\
XVII & 60 & 3 & 371 & 23\% & 377 & 54 \\
XVIII & 54 & 3 & 306 & 34\% & 418 & 54 \\
\bottomrule
\end{tabular}
\caption{Legislative activity: counts and median duration by legislature. Median duration is shown separately for ordinary and decree laws.}
\label{tab:leg_description_extended}
\end{table}



\smallskip
\noindent\textbf{Procedural analyses.}  
In order to showcase possible inquiries enabled by \dset{}, and the added value brought by them, 
we  investigate how parliamentary procedures have changed over time. 
We do this by comparing process maps from two temporally distant periods considered above: 
legislatures X and XI, and legislatures XVII and XVIII.
This choice maximizes the potential to observe structural differences in the lawmaking process, while ensuring comparability: 
the two periods saw the alternation of 6 governments. 

\begin{figure}[t]
    \centering
    \begin{subfigure}[t]{0.4\textwidth}
        \centering
        \includegraphics[width=\textwidth]{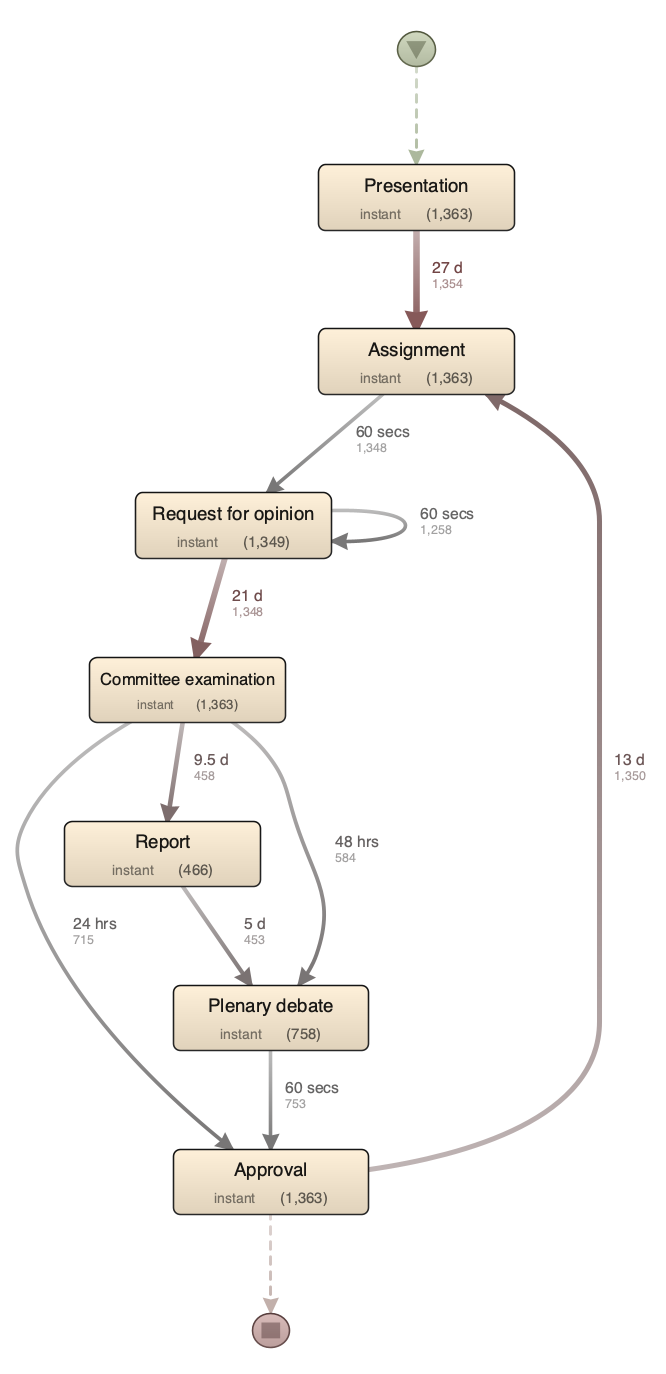}
        \vspace{-0.6cm}
        \caption{Maps for X and XI legislatures}
    \end{subfigure}
    \hfill
    \begin{subfigure}[t]{0.4\textwidth}
        \centering
        \includegraphics[width=\textwidth]{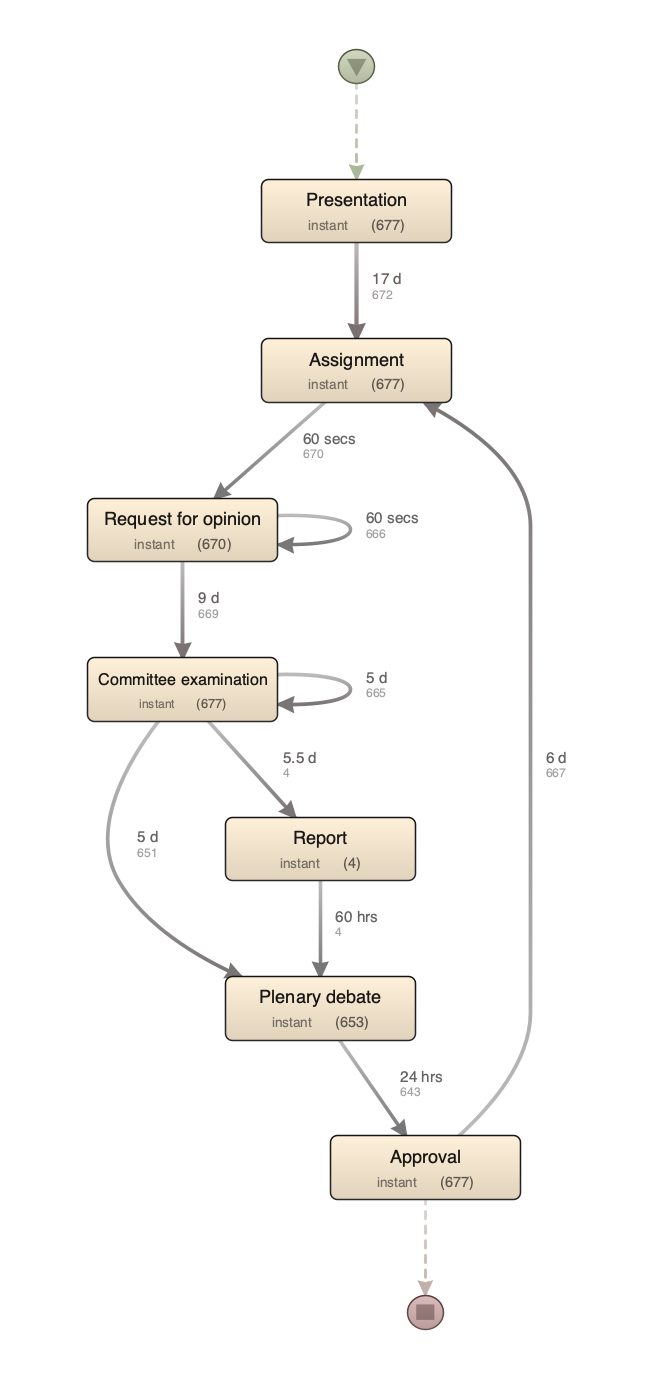}
        \vspace{-0.6cm}
        \caption{Maps for XVII and XVIII legislatures}
    \end{subfigure}
    \vspace{-0.1cm}
    \caption{
    Disco’s process maps showing median durations as primary metric, and case frequency as secondary metric (in parenthesis). We have set 100\% and
0\% as importance zoom for activities and paths, respectively. {
For each case\_id, we identified contemporary events, and incrementally adjusted their timestamps by one minute to preserve their order in the event log. 
}}
    \label{fig:processmaps}
\end{figure}

Figure~\ref{fig:processmaps} 
displays the resulting process maps. 
Disco allows setting \emph{importance filters} to focus on more frequent/important activities (the boxes) and paths (edges). In order to preserve all activities while focusing on only the most important relations, we have set 100\% and 0\% for activities and paths, respectively. 
These visualizations highlight significant changes in the legislative process over the decades. 
First of all, we observe a marked reduction in the time taken to complete certain activities (e.g., from \code{Presentation} to \code{Assignment}, or from \code{Request for opinion} to \code{Committee examination}). 
This seems to contrast with the discussion given for Table~\ref{tab:leg_description_extended}, where we found that the two most recent legislatures have (slightly) longer median durations.  
%
However, this can be explained by a notable change from a process-oriented perspective. In particular, we see a marked decline in the use of the so-called \textit{sede deliberante} procedure (the committee gets assigned deliberating powers), in which a committee directly approves a law without requiring debate and approval by the full assembly (\code{Plenary debate}). 
This change is clearly visible in the process maps: the mentioned path is present in Figure~\ref{fig:processmaps} (a)~\footnote{We remark that \code{Report} is (optionally) performed only before \code{Plenary debate}. Therefore, it can be skipped in the usual procedure, while it is always skipped with the sede deliberante procedure.}, while it is absent in Figure~\ref{fig:processmaps} (b). This is because, becoming less frequent, it is hidden by the low importance zoom used for paths. 
{In fact, while this mechanism accounted for 53\% of approved laws in the earlier legislatures (see secondary metric in the considered path in Figure~\ref{fig:processmaps} (a), counting 715 cases out of 1363, its usage dropped to just 16\% (651 laws out of 677 follow the `normal' paths outgoing from \code{Committee examination}) in the later legislatures.  
A deeper analysis of the reasons and impacts behind this change is beyond the scope of this paper. 
However, it may partly explain the longer enactment durations observed in Table~\ref{tab:leg_description_extended} for the last two legislatures, despite the shorter durations shown in Figure~\ref{fig:processmaps} (b). }


\clearpage
\section{
Download, loading and usage instructions, and
License specification.}


The dataset is freely available at the following Zenodo link~\cite{prolificzenodo} 
under license CC BY-NC 4.0. 
In addition, the working GitHub repository of \dset{} is available at \github, where it can be tested online on Google Colab without requiring any local installation (see \colab{}). 
The dataset comes with  
a notebook to load and manipulate the data, including pre-processing done before feeding the data into Fluxicon Disco to obtain the maps in this manuscript.

\section{Conclusion}
We have presented \dset, \dsetext, a curated event log dataset on the Italian lawmaking process from 1987 to 2022. The dataset has been created using an automated pipeline that extracts unstructured texts on the procedural history of Italian laws available in the Normattiva portal, and exploits large language models (LLMs) to structure them into events, allowing for process-oriented data science analyses.
Our work aligns with two recent trends in the process mining (PM) community aiming at applying PM to the legal context, and to integrate PM with LLMs.
We have presented the structure of \dset, as well as examples of novel analyses enabled by it. Despite our process-oriented analyses being preliminary, we discovered notable changes over years to the process of Italian lawmaking. 
The dataset is available at~\cite{prolificzenodo}. 

\paragraph{Acknowledgments.}
	Work partially supported by 
    SMaRT COnSTRUCT (CUP J53C24001460006), in the context of FAIR (PE0000013, CUP B53C22003630006) under the National Recovery and Resilience Plan (Mission 4, Component 2, Line of Investment 1.3) funded by EU - NextGenerationEU.

\bibliography{sample-ceur}




\end{document}